\documentclass[10pt,twocolumn,letterpaper]{article}

\usepackage[pagenumbers]{files/cvpr} 
\usepackage{color}

\definecolor{orange}{RGB}{255,107,0}

\newcommand{\bb}{\boldsymbol}

\usepackage{amsmath,amsfonts,bm}

\def\eqref#1{equation~\ref{#1}}

\def\1{\bm{1}}

\def\vx{{\bm{x}}}
\def\vy{{\bm{y}}}
\def\vz{{\bm{z}}}

\DeclareMathAlphabet{\mathsfit}{\encodingdefault}{\sfdefault}{m}{sl}
\SetMathAlphabet{\mathsfit}{bold}{\encodingdefault}{\sfdefault}{bx}{n}

\def\gL{{\mathcal{L}}}

\def\sL{{\mathbb{L}}}

\def\sR{{\mathbb{R}}}

\def\sX{{\mathbb{X}}}

\newcommand{\E}{\mathbb{E}}

\DeclareMathOperator*{\argmax}{arg\,max}
\DeclareMathOperator*{\argmin}{arg\,min}

\usepackage{graphicx}
\usepackage{amsmath}
\usepackage{amssymb}
\usepackage{booktabs}

\usepackage{url}
\usepackage{enumerate}
\usepackage{bbm}
\usepackage{graphics}
\usepackage{arydshln}
\usepackage{diagbox}
\usepackage{listings}
\usepackage[accsupp]{axessibility}  

\usepackage[pagebackref,breaklinks,colorlinks]{hyperref}

\usepackage[capitalize]{cleveref}
\crefname{section}{Sec.}{Secs.}
\Crefname{section}{Section}{Sections}
\Crefname{table}{Table}{Tables}     
\crefname{table}{Tab.}{Tabs.}

\def\cvprPaperID{1096} 
\def\confName{CVPR}
\def\confYear{2022}

\begin{document}
\setlength\abovedisplayskip{3pt}
\setlength\belowdisplayskip{3pt}

\title{Cycle-Consistent Counterfactuals by Latent Transformations}

\author{Saeed Khorram, Li Fuxin\\
Collaborative Robotics and Intelligent
Systems (CoRIS) Institute\\
Oregon State University\\
{\tt\small \{khorrams, lif\}@oregonstate.edu}
}
\maketitle

\begin{abstract}
{
CounterFactual (CF) visual explanations try to find images similar to the query image that change the decision of a vision system to a specified outcome. Existing methods either require inference-time optimization or joint training with a generative adversarial model which makes them time-consuming and difficult to use in practice. We propose a novel approach, Cycle-Consistent Counterfactuals by Latent Transformations (C3LT), which learns a latent transformation that automatically generates visual CFs by steering in the latent space of generative models. Our method uses cycle consistency between the query and CF latent representations which helps our training to find better solutions. C3LT can be easily plugged into any state-of-the-art pretrained generative network. This enables our method to generate high-quality and interpretable CF images at high resolution such as those in ImageNet. In addition to several established metrics for evaluating CF explanations, we introduce a novel metric tailored to assess the quality of the generated CF examples and validate the effectiveness of our method on an extensive set of experiments.
}
\end{abstract} 

\section{Introduction}
\vspace{-0.05in}

With convolutional neural networks (CNNs) revolutionizing the space of automatic visual recognition, there have been many approaches attempting to better explain the inner workings of CNNs, including attribution maps\cite{selvaraju2017grad, khorram2021igos++}, concept-based explanations\cite{ribeiro2016should, ghorbani2019towards}, rule-based explanations\cite{frosst2017distilling}, prototype-based explanations\cite{NEURIPS2019_adf7ee2d}, etc. However, when presented to humans, those kinds of explanations were not necessarily easy to process. Recently,  a substantial user study \cite{jeyakumar2020can} shows that  Grad-CAM\cite{selvaraju2017grad}, LIME superpixels\cite{ribeiro2016should}, etc., were not as informative to humans as simple nearest neighbors from the training set. 

Those findings suggest that humans prefer to see examples that are just similar to the natural images rather than heatmaps, superpixels, etc. and \textit{counterfactual (CF)  explanations}\cite{goyal2019counterfactual,wang2020scout,dhurandhar2018explanations,wachter2017counterfactual, moore2019explaining, mothilal2020explaining} might be more useful in helping humans to understand deep networks. CF explanations show humans examples that are similar to the explanation subject but deep networks predict them as a different category. Such explanations have also been advocated by social scientists~\cite{miller2019explanation,wachter2017counterfactual} as a preferred mode of explanation.

This paper mainly deals with CF explanations in the visual domain. CF explanation in the visual domain is more difficult to generate than categorical inputs where one can simply search for adversarial examples~\cite{wachter2017counterfactual, moore2019explaining, mothilal2020explaining}. Methods that directly optimize for perturbations in the input space \cite{dhurandhar2018explanations} often lead to adversarial solutions~\cite{szegedy2013}, which manipulate CNN predictions with imperceptible changes. Adversarial examples are usually off the data manifold, where CNNs are fooled because they do not generalize to the kinds of data that have never been seen in training. 
In addition, finding and replacing patches of features from images in the CF class to a query one \cite{goyal2019counterfactual} also moves the images off the natural image manifold by creating irregular edges.


Successful CF explanations usually \text{avoid} being adversarial by staying on the same data manifold the network has been trained on. Hence, prior work usually utilizes a generative model such as a generative adversarial network (GAN) or variational autoencoder (VAE) that ensures the generated CF example lies on the data manifold. For example, ExplainGAN~\cite{samangouei2018explaingan} jointly trains a GAN for each category along with a mask generator that generates a masked region from the latent code of the image, so that after the masked region is transformed the image would be classified as another category by the CNN. Some other algorithms\cite{Rodriguez_2021_ICCV} optimize for latent codes of a VAE model that will generate an image similar to the original one yet classified as another category.

Despite these prior work, it remains difficult to apply CF explanations in practice. One can consider two realistic use cases for explanation algorithms. The first is \textit{debugging}, where users attempt to check why CNN is making a certain wrong classification. The second is \textit{knowledge gathering}, where users may try to utilize explanations to understand subtle differences between two classes. In both cases, it would be beneficial for the user to quickly churn through many examples to help building their mental model. Even better, the user may want to make some realistic edits (e.g. based on GANs)) to the image and then obtain a new CF out of the edited image. 
In those cases, it would be ideal if CF images can be generated \textit{on-the-fly}. However, most previous approaches solve an optimization for each image \cite{dhurandhar2018explanations, goyal2019counterfactual, liu2019generative, russell2019efficient, mahajan2019preserving, mothilal2020explaining}, which often makes generating CF examples \textit{time-consuming}. 


In this paper, we propose a novel approach that optimizes for a nonlinear transformation in the latent space. The transformation morphs the latent code of an input image into a CF latent vector that can be decoded into an image which looks similar to the original one, but has semantically meaningful, perceptible differences so that CNN classifies it as another category. Different from \cite{samangouei2018explaingan}, our approach does not require joint training with GANs. It utilizes a pretrained generative model (GAN/VAE) hence can easily adapt to current and future generative algorithms that are being proposed every day due to significant ongoing research. As an example, this enables our framework to go beyond simple datasets and generate high-resolution images, e.g. ImageNet, with the current GAN algorithms available now. We further adopt a cycle-consistency loss function~\cite{zhu2017unpaired} that improves the consistency and performance of our approach.

Furthermore, we evaluate our approach comprehensively in a quantitative manner. For CF explanations, literature has suggested certain properties to be desirable~\cite{moraffah2020causal, verma2020counterfactual}:
\begin{enumerate}[I]
    \itemsep-0.25em 
    \item \text{\sc Validity.} The model should assign the CF examples $\vx'$ to the CF class $c'$ in order to be valid. \label{pr:val}
    
    \item \text{\sc proximity.} The CF examples $\vx'$ should stay as close as possible (in terms of some distance function) to the original query instance $\vx$. \label{pr:prx}
    
    \item \text{\sc Sparsity.} Minimal number of query features should be perturbed in order to generate CF examples. \label{pr:spr}
    
    \item \text{\sc Realism.} The CF examples should lie close to the data manifold so that it appears realistic. \label{pr:real}
    
    \item \text{\sc Speed.} The CF explanations should be generated in interactive speed in order to be deployed in real-world applications. \label{pr:speed}
    \end{enumerate}
\vspace{-0.05in}
For example, adversarial examples may be valid CFs, but they fail in terms of realism. 
We propose to use a set of metrics that comprehensively measure all these aspects, including a novel metric that inspects the quality of the CF examples across a series of changes. 


Below we list our contributions in this paper:

    
    
    

\vspace{-0.05in}
\begin{itemize}
    \itemsep-0.25em 
    \item We introduce a novel framework to generate realistic CFs at high resolution by learning a transformation in the latent space of a pretrained generative model.
    
    
    
    \item We propose a set of novel quantitative evaluation metrics tailored for counterfactual explanations.
    
    \item Extensive qualitative and quantitative evaluations show the effectiveness of our method and its capability to generate high-resolution CF images by plugging into existing generative algorithms.
\end{itemize}
\vspace{-0.05in}

\vspace{-0.125in}
\section{Related Work}
\vspace{-0.05in}

\textbf{Counterfactual Visual Explanation.}
While many of the previous approaches in CF explanation focus on categorical data \cite{wachter2017counterfactual, moore2019explaining, mahajan2019preserving, mothilal2020explaining, pawelczyk2020learning}, in this paper, we mainly concentrate on generating CF examples in the vision domain. One of the early approaches on counterfactual \textit{visual} explanation is \cite{goyal2019counterfactual} where CFs are generated by exhaustively searching for feature replacement between the latent feature of query and CF images. 
Due to the exhaustive search for individual samples, this method is slow in practice and the generated CF images are oftentimes off the data manifold. Later, \cite{wang2020scout} proposed SCOUT in which the regions that are exclusively informative for the query or the CF classes are discovered using attribution maps. However, this work does not compose CF images and the explanations are limited to highlighting regions over images. Unlike our work, the quality of the explanations on both aforementioned approaches relies on the choice of CF images from the training set and the heuristics used for finding them. 

\cite{dhurandhar2018explanations} proposes a contrastive explanation framework with the goal of finding minimal and sufficient input features in order to justify the prediction or finding minimal and sufficient perturbations in order to change the classifier's prediction from the query class to a CF one (pertinent negative). Applying such perturbations is limited to gray-scale images. Although the authors suggested using an auto-encoder loss term to align CF examples to the distribution of the original data, the generated CF examples are adversarial and off the data manifold. The authors in \cite{looveren2021interpretable} proposed to incorporate a prototype loss in the optimization of \cite{dhurandhar2018explanations}, making the generated CFs more interpretable. These methods usually push the generated images off the manifold of the natural images and are limited to simple datasets. We have also observed that occasionally their optimizations do not converge.

ExplainGAN \cite{samangouei2018explaingan} composes CF examples by filling a masked area over the input using a generator. Their design has an additional mask generator that needs to be trained jointly with the GAN. This is an impediment to plugging their method into existing GANs and extending the scope of their work to complicated datasets such as ImageNet. 

More recently, \cite{sauer2020counterfactual} decomposes image generation into parallel mechanisms (shape, texture, and background) and the distributions over the individual mechanisms are learned. 
Their work generates high-resolution images, but 
do not explain the decision of a classifier. 

\cite{poyiadzi2020face} builds a graph over all the candidates in the training set and selected CFs from it to respect the underlying data distribution. This assumes a counterfactual example to the query image can be found among the training examples, which may not be always true. \cite{Rodriguez_2021_ICCV, joshi2019towards, goyal2019explaining} use conditional VAE-based architectures to generate CFs. They solve individual optimizations for each sample. Similarly, many other previous CF explanation methods \cite{dhurandhar2018explanations, goyal2019counterfactual, liu2019generative, russell2019efficient, mahajan2019preserving, mothilal2020explaining} have separate optimizations for each query image. This obstructs their applications in real-time. In turn, our method learns a transformation from query to CF (and vice versa) over the course of training. At inference time, there is no optimization to be solved and our method is suitable for interactive use.

\cite{zhu2017unpaired} learns an unpaired image-to-image translation using cycle-consistent adversarial training. However, it has a different goal than ours. While our method can explain the decisions from \textit{any} classifier, their method does not and instead uses two separate discriminators in order for the transformed images to lie in the target image class.

\textbf{ Latent Manipulations in Generative Models}
It has been shown that GANs learn interpretable directions in their latent space and meaningful changes can be obtained by steering in such directions. \cite{jahanian2019steerability} shows that by linearly walking in the latent space of pretrained GANs, simple edits on images (e.g. zoom, rotation, etc.) can be learned. \cite{gu2019mask,shen2020interpreting,yang2021discovering} aim to learn interpretable directions in the latent space of the GANs for attribute manipulation such as face editing (e.g. age, expressions, etc.). However, their manipulations do not explain the decision of an external classifier. Our method, on the other hand, explains any given classifier using the same GAN/VAE backbone.

\vspace{-0.05in}
\section{Methodology} \label{sec:method}

\subsection{Generating CFs by Transformation in the Latent Space}\label{sub:cfl_steer}

As some prior work \cite{samangouei2018explaingan,Rodriguez_2021_ICCV}, we utilize a generative model in order to obtain more realistic counterfactual examples that stay close to the data manifold. Toward that goal, we follow the recent idea of steerability in the latent space of generative models \cite{jahanian2019steerability} and propose to \textit{learn} a transformation in the latent space to obtain the CFs.
Given a pretrained classifier $f$ that we are attempting to explain, a pretrained generator $G$, an input (query) image $\vx \in \sX_c$ from the images in the training set with the query class $c$, and a target CF class $c'$, we re-define the CF generation problem \cite{wachter2017counterfactual, joshi2018xgems} to learn a (non-linear) transformation $g: \sR^D \xrightarrow{} \sR^D$ in the latent space of the generator that maps the latent code of the input ($\vz_x$) to a CF one,
\begin{align} \label{eq:cfl_g}
    g^* = \argmin_{g}\;&\E_{\vx \in \sX_c} \left[\gL_{cls}\left(f\left(\vx'\right), c'\right) + \gL_{prx}\left(\vx',\vx\right)\right] \notag \\
    s.t. \quad &\vx' = G\left(g^n\left(\vz_x\right)\right), \;\; \vz_x=E\left(\vx\right)
\end{align} 
where $\vx'$ is the generated CF and $g^n(.)$ is an $n$-th order function decomposition $g(g(g(\dots)))$ --- $g$ is recursively applied $n$ times, mimicking discrete Euler ODE approximations. Here, $g(.)$ is estimated using a simple neural network. $\gL_{cls}$ is the \textit{classification} loss so that the generated CF $\vx'$ belongs to class $c'$ and $\gL_{prx}$ is the \textit{proximity} loss encouraging $\vx'$ to be proximal to the input $\vx$. To get to the latent code $\vz_x$ from the query image $\vx$, a pretrained encoder $E: \sR^{C\times H\times W} \xrightarrow{} \sR^D$ can be used. 

It is worth mentioning that the main difference between this formulation and prior explanation  work~\cite{joshi2018xgems,Rodriguez_2021_ICCV} is that $g$ is a transformation that can be directly applied on any new query image once learned, whereas prior work would need to solve separate optimization problems for every new image. The difference between this formulation and a regular conditional GAN is that our approach is used to explain a generic classifier $f$ that is independent of the GAN, whereas conditional GANs use their discriminators to encode class knowledge. In practice, a significant amount of work is put into training discriminative classifiers and it would be desirable to have a tool that can diagnose any pre-trained classifier with a joint re-training with the GAN.

Note that in our formulation $E$ is not an integral part and $\vz_x$ can be obtained by directly sampling from the latent space distribution of the generator
. To put it differently, for the purpose of training $g$, our method does \textit{not require} access to the images $\vx$ and sampling $\vz_x$ is sufficient. When directly sampling $\vz_x$, the input image is $\vx = G(\vz_x)$. For unconditional generative models, rejection sampling needs to be used to select $\vz_x$ --- based on the classifier's prediction $\vz_x = \{\vz|c=\argmax_{}f(G(\vz))\}$. In the case of conditional generative models, sampling $\vz$ from class $c$ is trivial. Directly sampling $\vz_x$ is particularly advantageous when using GANs as obtaining the latent GAN codes from images is still an open research topic \cite{xia2021gan}. When using VAEs as the generative models, however, obtaining latent code is straightforward where $E$ is the encoder of the VAE. 


\begin{figure*}[t!]
\begin{center}
{ 
\includegraphics[width=0.85\linewidth]{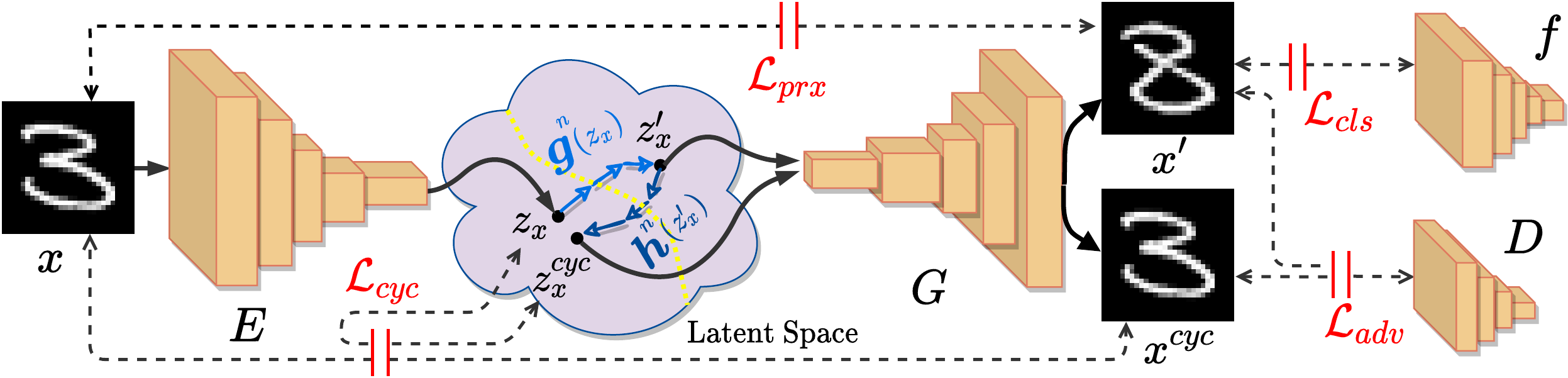}
}
\end{center}
\vskip -0.15in
\caption{\small \textbf{Cycle-Consistent Counterfactuals by Latent Transformations (C3LT)}. This figure illustrates the architecture of our proposed framework. After a latent code $\vz_x$ is obtained, our method transformed it to the CF latent code $\vz_x'$ using $g$. The CF example can be obtained by $\vx' = G\left(g^n\left(\vz_x\right)\right)$. The inputs to the loss functions are outlined via dashed lines. The classifier $f$ and discriminator $D$ are only used during the training. (Best viewed in color)}\label{fig:cfl}
\vskip -0.2in
\end{figure*}

\subsection{From Query to CF and Back: Cycle-Consistent CF Generation}\label{sub:cfl_cyc}
Finding a transformation $g$ is highly under-constrained and there might be many solutions to the optimization problem (\ref{eq:cfl_g}) that satisfy the CF properties equally. To regularize the optimization, we modify the objective and incorporate \textit{cycle consistency} \cite{zhu2017unpaired} between the query and CF latent codes. That can be achieved by introducing another transformation $h: \sR^D \xrightarrow{} \sR^D$ that estimates the inverse of $g$, \ie, finds a (non-linear) trajectory in the latent space that maps the CF latent code back to the query one, \ie, $\vz_x \approx \vz_x^{cyc}$ where $\vz_x^{cyc}= h^n \left(g^n\left(\vz_x\right)\right)$. We also define the \textit{cycled} query image as $\vx^{cyc}=G\left(\vz_x^{cyc}\right)$ and add the \textit{cycle} loss to objective (\ref{eq:cfl_g}). Note that $\vx'$ and $\vx^{cyc}$ belong to two distinct classes (the CF and the query class, respectively). 

Although generating CFs from the latent space of generative models helps with staying close to the data manifold, it does not guarantee such. To ensure staying on the data manifold, we add an \textit{adversarial} loss to objective (\ref{eq:cfl_g}). More formal descriptions of the adversarial and cycle losses are presented later in this subsection.

Here, we formalize the main objective of our method, Cycle-Consistent Counterfactuals by Latent Transfomrations (C3LT). For cycle consistency, our method requires access to samples from both the query and the CF classes. Given an image $\vx \in \sX_c$ from the images in the training set with the query class $c$, an image $\vy \in \sX_{c'}$ from the CF class $c'$, our method learns transformations $g^*$ and $h^*$,
\begin{align} \label{eq:cfl}
    g^*, h^* = \argmin_{g, h}\;&\E_{\vx \in \sX_{c}} \left[\gL_{c3lt}(\vx,c',g,h)\right] + \notag \\
    &\E_{\vy \in \sX_{c'}}\left[\gL_{c3lt}(\vy,c,h,g)\right] 
\end{align}
where,
\begin{align}
    \gL_{c3lt}(\vx,c',g,h) =& \gL_{cls}\left(f\left(\vx'\right), c'\right) + \gL_{prx}\left(\vx',\vx\right) +\notag \\ 
    & \gL_{cyc}\left(\vx^{cyc}, \vx\right) + \gL_{adv}(\vx', \vx^{cyc}) \notag \\
    s.t. \quad \vx' = G\left(\vz_x'\right),&\;\; \vz_x'=g^n\left(\vz_x\right)\notag,
    \vx^{cyc}=G\left(\vz_x^{cyc}\right),\\\quad \vz_x^{cyc}=h^n \left(\vz_x'\right) , &\; \vz_x=E(\vx),
    \label{eq:cfl_loss}
\end{align}

C3LT learns transformations between the query and CF classes at the same time, hence the query and CF notations are interchangeable. For brevity purposes, we skip the formal definition of the $\gL_{c3lt}(\vy,c,h,g)$. Fig. \ref{fig:cfl} shows the architecture of our proposed framework. In what follows, we define the individual loss terms in Eq. (\ref{eq:cfl_loss}):

\textbf{Classification Loss $\gL_{cls}$} encourages the generated CF examples to 
be classified as the CF class. 
We use the Negative Log-Likelihood loss,
\begin{equation}\label{eq:cls}
    \gL_{cls} = -\log\left(f_{c'}\left(\vx'\right)\right)
\end{equation}
where $f_{c'}\left(\vx'\right)_{c'}$ is the output of the classifier for class $c'$.

\textbf{Proximity Loss $\gL_{prx}$} helps the generated CF examples to stay close to the query image in terms of some distance function \ie CFs that are proximal to the query ones. It is also desirable that the CFs have sparse changes compared to the input images \ie only a few input features change. To that end, we opt to choose an $L1$ loss term for the proximity loss. In addition, we use entropy and smoothness losses ($\gL_{entr}$ and $\gL_{smth}$)\cite{samangouei2018explaingan} over the absolute difference between the query and CF images to encourage changes to be more sparse and local,
\begin{align}\label{eq:prx}
    \gL_{prx} = ||\vx - \vx'||_1 + \gL_{entr}\left(\vx,\vx'\right) + \gL_{smth}\left(\vx,\vx'\right)
\end{align}

\textbf{Cycle-Consistency Loss $\gL_{cyc}$} enforces the cycle-consistency between the latent codes for the query and CF classes. In addition to the latent codes, we use Perceptual similarity \cite{johnson2016perceptual} over the input $\vx$ and the cycled image $\vx^{cyc}$,
\begin{align}\label{eq:cyc}
    \gL_{cyc} = &\sum_{l\in \sL}||\hat{f}^{l}(\vx^{cyc})-\hat{f}^{l}\left(\vx\right)||_1 + ||\vx^{cyc}-\vx||_1 \notag \\
    &+||\vz_x - \vz_x^{cyc} ||_1 \\
    s.t.\quad & \vz_x^{cyc} = h^n \left( g^n\left(\vz_x\right)\right), \;\; \vx^{cyc}=G\left(\vz_x^{cyc}\right) \notag
\end{align}
where $\hat{f}^{l}(\vx)$ is the intermediate features of a pretrained classifier $\hat{f}$ at layer $l$ for a given input $\vx$, and $\sL$ is the set of all target layers. Note, classifier $\hat{f}$ can be the same as or different from the original classifier $f$ that we are explaining.

\textbf{Adversarial Loss $\gL_{adv}$} helps the generated CFs and \textit{cycled} images to lie close to the manifold of the original data using the discriminator $D$,
\begin{align}\label{eq:adv}
    \gL_{adv} = & \log\left(1-D\left(\vx^{cyc}\right)\right)+\log\left(1-D\left(\vx'\right)\right)
\end{align}
so that $\vx^{cyc}$ and $\vx'$ pose to the discriminator as real images.

\vspace{-0.05in}
\subsection{Inference} \label{sub:inference}
At the inference time, when an encoder $E$ is available, the input $\vx$ goes through the encoder to obtain the latent code $\vz_x=E(\vx)$. It is then transformed by $g^*$, followed by passing through the generator $G$ to obtain the CF example $\vx' = G\left(\left(g^*\right)^n\left(\vz_x\right)\right)$. This results in \textit{fast} inference and makes our method suitable for interactive applications --- unlike many of the previous approaches \cite{dhurandhar2018explanations, goyal2019counterfactual, liu2019generative, russell2019efficient, mahajan2019preserving, mothilal2020explaining, joshi2018xgems, Rodriguez_2021_ICCV} where the CFs are generated by solving optimization problem for individual inputs. 

When an encoder $E$ is not available, the inference is slightly different; given an input image $\vx$, the latent code is calculated by $\vz_x^* = \argmin_{\vz} \gL_{}(G(\vz), \vx)$ \cite{xia2021gan,gu2020image} which is slower than when the encoder is available. The rest is similar to the above procedure. Note that when there are no input images, the query and CF classes can be inspected by sampling $\vz_x$ directly from the latent space distribution.

\vspace{-0.05in}
\section{Experiments} \label{sec:exp}

\subsection{Setup} \label{sub:setup}

\textbf{MNIST and Fashion-MNIST.}
We evaluate the C3LT method against CF explanation baselines, namely, Contrastive Explanation Method (CEM) \cite{dhurandhar2018explanations}, Counterfactual Visual Explanation (CVE) \cite{goyal2019counterfactual}, and ExplainGAN (ExpGAN) \cite{samangouei2018explaingan} on the MNIST \cite{lecun1998gradient} and Fashion-MNIST \cite{xiao2017} datasets by both qualitative inspection and an extensive set of quantitative metrics. Due to the similarities in CF explanation and adversarial attacks, we also generate adversarial examples on the query images using the PGD attack (denoted as Adv. Attack) \cite{madry2018towards} with the CF class as the target. Images from both datasets have $28 \times 28$ resolutions and 10 classes. We use the standard train/test split. The C3LT and \cite{samangouei2018explaingan} use the examples from the query and CF classes in the train set ($\sim$6,000 samples/class) for the purpose of training and the examples from the query/CF class in the test set ($\sim$1,000 samples/class) for evaluation. While we used official implementations for \cite{dhurandhar2018explanations, goyal2019counterfactual}, we could not find any available implementations for \cite{samangouei2018explaingan} and implemented it ourselves.

\cite{dhurandhar2018explanations, goyal2019counterfactual} directly evaluate on the test set since they solve optimization problems for individual samples without any training. In addition, given an input $\vx$ from class $c$, \cite{dhurandhar2018explanations, goyal2019counterfactual} do not take user-specified CF class $c'$ as the target and just aim to change the classifier's output to the \textit{maximum-non-query} class $\argmax_{i\neq c} f(\vx)$. To have a fair comparison across all baselines, however, we slightly modify their objective and instead select the CF class $c'$ as the target. 

Similar to \cite{samangouei2018explaingan}, for the MNIST dataset, we use query and CF class pairs (3, 8), (4, 9), and (5, 6). For Fashion-MNIST, we use (coat, shirt), (t-shirt, pullover), and (sneaker, boot). Unlike C3LT and \cite{samangouei2018explaingan}, \cite{dhurandhar2018explanations, goyal2019counterfactual} do not guarantee a solution, and we found their optimizations occasionally do not converge hence would not be able to find any CF explanations. For a fair comparison, we only consider the samples that all methods successfully generated CF explanation for. The reported numbers in the following sections are averaged over all samples and pairs for each dataset. Across all methods, we use the same classifier $f$ for explaining.  The architecture of the classifier used is the same for both datasets where it obtains $99.4\%$ and $91.5\%$ test set accuracy on MNIST and Fashion-MNIST datasets, respectively.

Regarding C3LT, we train an encoder $E$ to map the input images to the corresponding code in the latent space of the generator. For the choice of the generator $G$, we used a pre-trained DC-GAN \cite{radford2015unsupervised} and PGAN \cite{karras2018progressive} for MNIST and Fashion-MNIST, respectively. We use similar discriminators as used by DCGAN method. Moreover, we used a simple 2-layer fully-connected neural network with ReLU activation for the choice of transformations $g$ and $h$.

\textbf{ImageNet from BigGAN.}
To showcase the capability of our framework in generating CFs on high-resolution real-world data, we use C3LT on ImageNet \cite{deng2009imagenet}-trained BigGAN\cite{brock2018large}, a conditional GAN that generates high-fidelity and high-quality images. To the best of our knowledge, the C3LT is the first CF explanation method to generate CFs that explain classifiers for high-resolution natural images such as ImageNet. This is possible due to the flexibility of our framework and its modular nature. Here, we sample $\vx_x$ directly from the latent space distribution $\vz_x \sim \mathcal{N}(0,I)$ with truncation $0.4$ and an encoder $E$ is not required. We opt to use pre-trained \textit{BigGAN-deep} at $256 \times 256$ resolution. We use the (\textit{leopard, tiger}), (\textit{Egyptian cat, Persian cat}), (\textit{rooster, hen}), (\textit{husky, wolf}), and (\textit{pembroke corgi, cardigan corgi}) class pairs. The experiments on BigGAN are limited to C3LT (our method) as baselines are incompetent in generating meaningful CFs and we do not quantitatively compare against them. For further details on the experiments, please refer to the supplementary materials \ref{sec:hyperparams}.


\begin{figure*}[hbt!]
\begin{center}
{ 
\includegraphics[width=0.9 \linewidth]{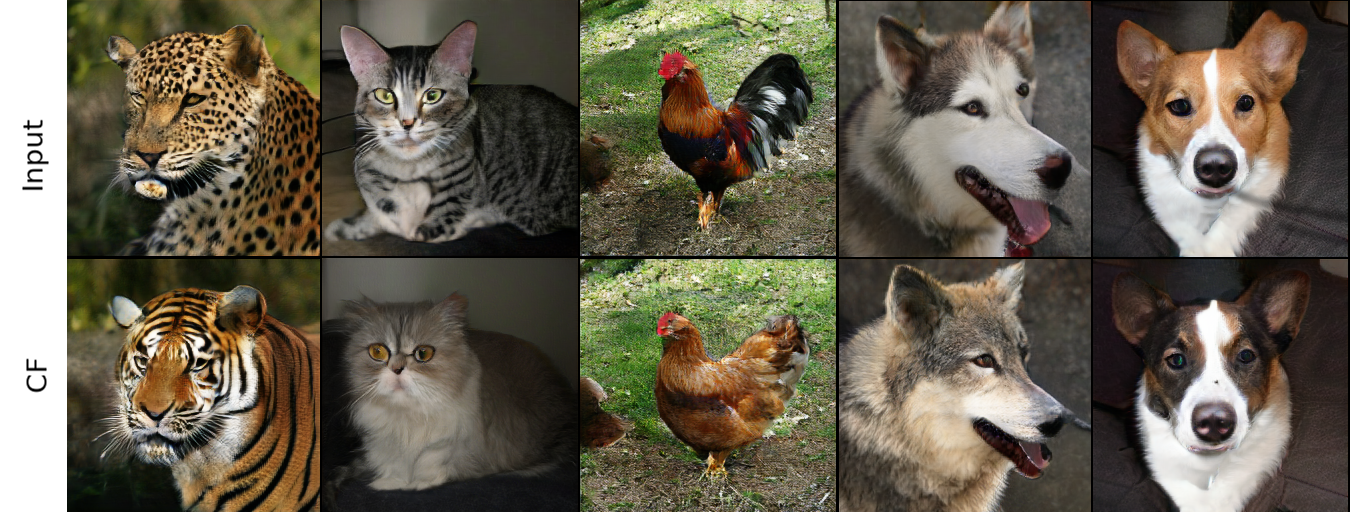} 
}
\end{center}
\vskip -0.15in
    \caption{\small \textbf{High-resolution ($256 \times 256$) Counterfactual Generation}. This figure shows high-resolution CF explanations generated using C3LT for (\textit{leopard, tiger}), (\textit{Egyptian cat, Persian cat}), (\textit{rooster, hen}), (\textit{husky, wolf}), and (\textit{pembroke corgi, cardigan corgi}) CF pairs, respectively from left to right.} \label{fig:biggan}
\vskip -0.15in
\end{figure*}

\begin{figure}[thb!]
\begin{center}
{ 
\includegraphics[width=0.9\linewidth]{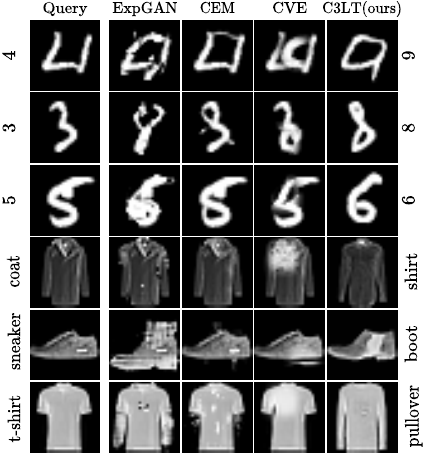} 
}
\end{center}
\vskip -0.15in
\caption{\small \textbf{Qualitative Comparison of the CFs}. This figure shows CF explanations obtained from our method (C3LT) and baselines across MNIST and Fashion-MNIST datasets. Broadly, we find the generated CFs from the CEM and CVE to be adversarial and off the data manifold. Compared to the ExpGAN, the generated CFs from our method are consistently more realistic and interpretable.} \label{fig:vis_nist}
\vskip -0.15in
\end{figure}

\vspace{-0.05in}
\subsection{Qualitative Inspection of the Counterfactuals} \label{sub:exp_qual}
Fig. \ref{fig:biggan} shows some examples of generated CFs for high-resolution images using BigGAN. This showcases that our method can be plugged into state-of-the-art GAN models and generate CFs by finding transformations in their latent space. It can be observed that our method pays attention to both \textit{foreground} and \textit{background} in the image and mainly keeps the background the same. C3LT found transformations in both the \textit{global shape} and the \textit{texture} of the objects according to their category. For instance, for going from rooster to hen, the shape and texture are changed, e.g., smaller legs, smaller comb, and bigger belly, and the texture of the breast is slightly altered. For the leopard to tiger and corgi examples, however, the main transformations are occurring in the texture and color of the objects. 

In Fig. \ref{fig:vis_nist}, we present the CF examples obtained from C3LT and other baselines from the MNIST and Fashion-MNIST datasets. We show comparison across all the pairs used in the evaluation. We found ExpGAN to generate more interpretable explanations than other baselines. However, the CFs are often un-natural with diffuse perturbations (e.g. images 4 and \textit{t-shirt}). We suspect this is due to their CF composition mechanism using a mask over the input. Further, we found the CFs to be occasionally adversarial where the mask generation fails (e.g. images 3 and \textit{coat}). Not to our surprise, the CFs obtained from the CEM were mostly adversarial and the perturbations were hardly perceptible (e.g. images 5 and \textit{coat}, \textit{sneaker}). Although replacement of patches of pixels and losing the global shape might fool the CNN \cite{brendel2018approximating}, we did not find the CFs from the CVE to be interpretable (e.g. images 5 and \textit{coat}). Inspecting through the generated CFs, we found our method to consistently generate interpretable and realistic images. Quantitative results obtained from section \ref{sub:real} support our findings.

\vspace{-0.1in}
\subsubsection{Can C3LT Generate CFs on Nonsimilar Classes?}
\begin{figure}[h]
\vskip -0.1in
  \centering
  \includegraphics[width=0.9\linewidth]{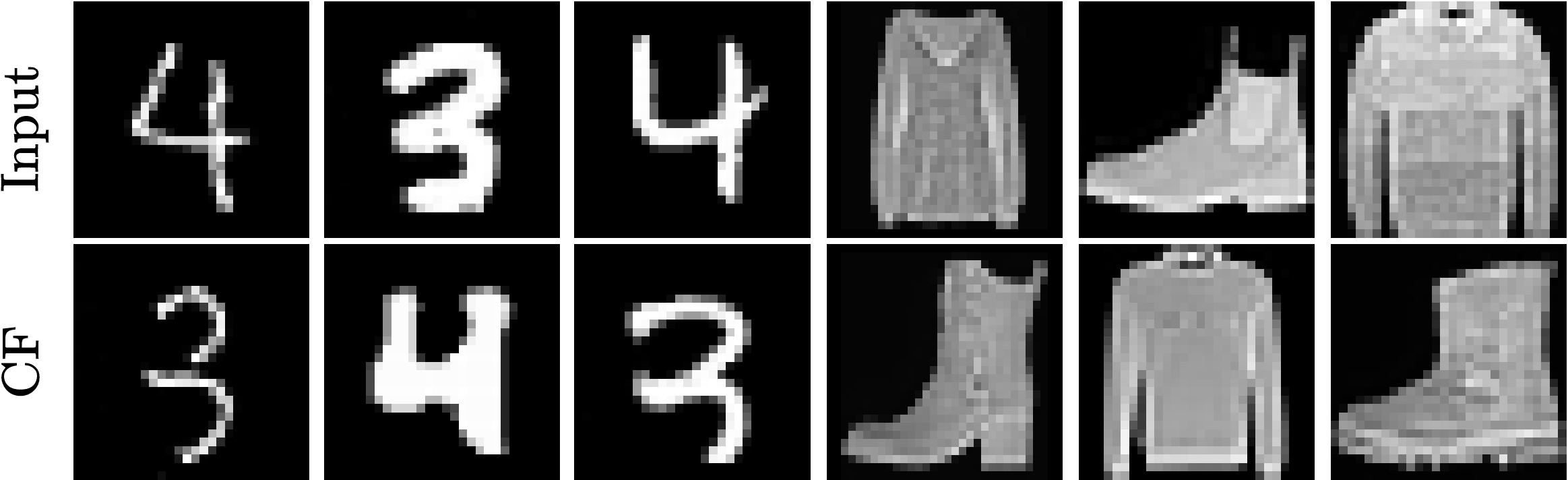}
  \vskip -0.05in
  \caption{CF examples on non-similar classes using C3LT.} 
  \label{fig:dist}
\end{figure}

\vspace{-0.05in}
Although the CF classes in the experiments were chosen to be close to the query one, C3LT, as a method, generates CF examples for any class pairs. To show this, we generate CFs for non-similar class pairs ($3$, $4$) from MNIST and (boot, pullover) from Fashion-MNIST in Fig. \ref{fig:dist}. Note that although the changes are significant, certain latent attributes that are independent of the category are still preserved, e.g., in the pairs between handwritten $3$s and $4$s, the stroke width and writing style are preserved, and in the boots/pullover case, one can see that slim clothes transfer to slim boots and vice versa.



\begin{table*}[ht!]
\caption{\small \textbf{COUnterfactual Transition (COUT) Scores}. The COUT metric measures the quality of the generated CFs and implicitly evaluates their {validity} and {sparsity} properties (\ref{pr:val},\ref{pr:spr}), \ie, it favors \textit{small} amount of changes that \textit{maximizes} the output score for the CF class (minimizes the output score for the query class). COUT is computed as the difference between the area under the curve for the classifier's output of both the query ($\textit{AUPC}_c$) and CF ($\textit{AUPC}_{c'}$) classes. The higher the COUT score the better.}
\label{table:cout}
\begin{center}
\vskip -0.2in
\begin{small}
\resizebox{0.8\textwidth}{!}{%

\begin{tabular}{l|cc|cc|cc|cc||cc}

{Methods} & 
\multicolumn{2}{c|}{ExpGAN \cite{samangouei2018explaingan}} &
\multicolumn{2}{c|}{CEM\cite{dhurandhar2018explanations}}  &
\multicolumn{2}{c|}{CVE \cite{goyal2019counterfactual}} & 
\multicolumn{2}{c||}{C3LT (ours)} & 
\multicolumn{2}{c}{Adv. Attack \cite{madry2018towards}}\\

\cline{1-11}  
{}  &   Mnist&FMnist &  Mnist&FMnist  &  Mnist&FMnist  &  Mnist&FMnist  &  Mnist&FMnist\\ 
\cline{2-11}
{$AUPC_{c'}\uparrow$}  &  {$0.967$}&{$0.920$}  &  {$0.301$}&{$0.347$}  &  {$0.209$}&{$0.275$}  &  {$\bb{0.980}$}&{$\bb{0.958}$}  &  {$0.737$}&{$0.732$}\\ 

{$AUPC_{c}\downarrow$} &  {$0.040$}&{$0.062$}  &  {$0.555$}&{$0.427$}  &  {$0.767$}&{$0.638$}  &  {$\bb{0.031}$}&{$\bb{0.052}$}  &  {$0.266$}&{$0.481$}\\ 

\cline{2-11}
{$COUT\uparrow$}       &  {$0.927$}&{$0.858$}  &  {$-0.253$}&{$-0.080$}  &  {$-0.557$}&{$-0.363$}  &  {$\bb{0.948}$}&{$\bb{0.906}$}  &  {$0.471$}&{$0.251$}
\end{tabular}
}
\end{small}
\end{center}
\vskip -0.15in
\end{table*}
\vspace{-0.1in}
\subsection{Quantitative Evaluation of the CF Explanations} \label{sub:exp_quant}

\subsubsection{Counterfactual Transition Metric} \label{sub:cout}
One of the main challenges in explaining deep networks is defining automatic metrics for quantitative evaluation. The authors in ExpGAN \cite{samangouei2018explaingan} treat the mask generated from their method as a pixel-wise attribution map and evaluate it against attribution map baselines. However, we argue this would be a relevant comparison for \textit{factual} explanations rather than for CF ones. This is mainly due to the fact that in the evaluation of attribution maps, only changes in the output score for the query class are considered while changes in the output score of the CF class are dismissed. In the following, we propose a new metric called \textit{COUnterfactual Transition (COUT)} metric to address this shortcoming. 

\begin{figure}[tb!]
\begin{center}
{ 
\includegraphics[width=0.85\linewidth]{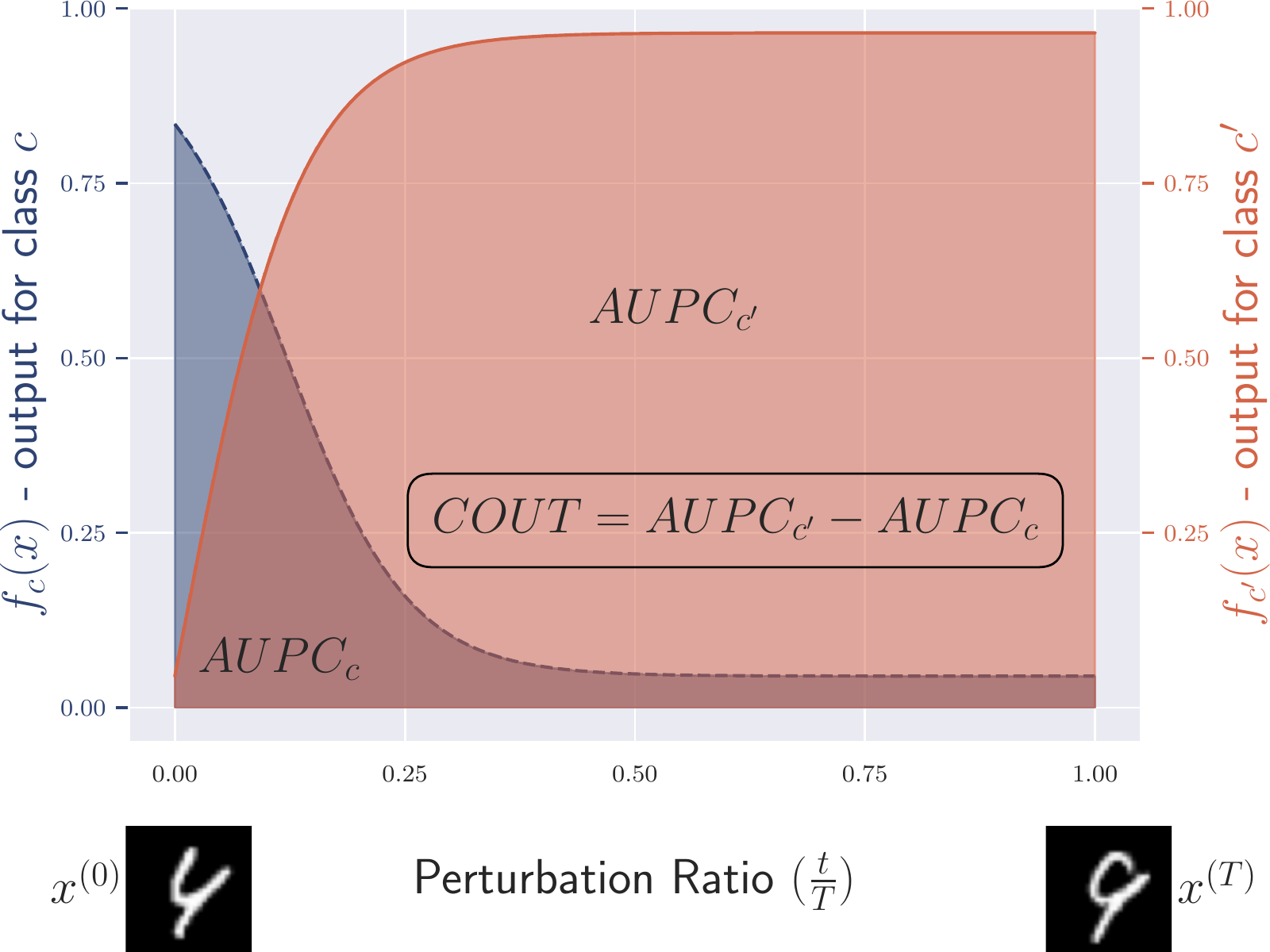} 
}
\end{center}
\vskip -0.15in
\caption{\small \textbf{COUnterfactual Transition (COUT) metric}. Here, we illustrate how the COUT metric is calculated. Starting from the query image $x^{(0)}$ (e.g. digit 4), after $T$ steps perturbations we get to the generated CF image $x^{(T)}$ (e.g. digit 9). The area under the curve for the classifier's output of both the query ($\textit{AUPC}_c$) and CF ($\textit{AUPC}_{c'}$) classes is calculated --- averaged over all evaluation data. The COUT metric is simply their difference.} \label{fig:cout}
\vskip -0.2in
\end{figure}

Inspired from the \textit{deletion} metric \cite{Petsiuk2018rise}, we devise a new metric to consider the changes in the output of the classifier for the query and the CF classes simultaneously, making it suitable for automatic evaluation of CF explanation methods. Given a query image $\vx$, a generated CF example $\vx'$, and a $mask \in [0,1]$ indicating the spatial location and \textit{relative} amount of changes over the query image needed to get to the CF one, the COUT metric is calculated as following; first, the pixel values in the (normalized) mask are sorted based on their values. Next, for a fix number of steps $T$, batches of pixels are inserted from the CF example into the query one according to the ordered masks values. The changes in the output score of the classifier for \textit{both} the query $c$ and CF $c'$ classes are measured. The Area Under the Perturbation Curve (\textit{AUPC} $\in [0,1]$) for each class $k \in \{c, c'\}$ is then calculated. Their difference is reported as the COUT $\in [-1,1]$ score,
\begin{align} \label{eq:tr_metric}
    &\text{COUT} = \textit{AUPC}_{c'} - \textit{AUPC}_{c} \\
    &\textit{AUPC}_{k} = \frac{1}{T} \left<
    \sum_{t=0}^{T-1} 
    \frac{1}{2}\left(f_{k}\left(\vx^{\left(t\right)}\right) + f_{k}\left(\vx^{\left(t+1\right)}\right)\right)
    \right>_{p_{\text{data}}} \notag
\end{align}
\noindent where $\vx^{\left(t\right)}$ is the input after $t \in \{0,\dots,T \}$ steps perturbations while $\vx^{(0)}$ being the query image ($\vx = \vx^{(0)}$) and the $\vx^{(T)}$ the CF one ($\vx'=\vx^{(T)}$) (see Fig. \ref{fig:cout}), $f_k()$ is the classifier's output for class $k$, and $<.>_{p_{\text{data}}}$ denotes the average over all images in the evaluation data.
Some methods such as ExpGAN explicitly generate the $mask$. However, for the rest of the baselines, given the CF and query images, it can be obtained by calculating the absolute difference between the images and normalizing it between 0 and 1. 

\begin{table*}[t!]
\caption{\small \textbf{Realism Comparison of the CFs}. In this table, we compare the generated CFs obtained from C3LT (our method) and its baselines in terms of realism, \ie, how close the generated CFs lie to the data manifold. Across all metrics, C3LT outperforms the baselines and generates high-quality CFs.}
\label{table:real}
\begin{center}
\vskip -0.2in

\begin{small}
\resizebox{0.93\textwidth}{!}{%
\begin{tabular}{l||cc|cc|cc|cc||cc|cc}

{Methods} & 
\multicolumn{2}{c||}{ExpGAN\cite{samangouei2018explaingan}} & 
\multicolumn{2}{c|}{CEM\cite{dhurandhar2018explanations}} &
\multicolumn{2}{c|}{CVE\cite{goyal2019counterfactual}}& 
\multicolumn{2}{c||}{C3LT (ours)}&
\multicolumn{2}{c|}{Original Data}& 
\multicolumn{2}{c}{Adv. Attack \cite{madry2018towards}}\\

\cline{2-13}
{}                 &  Mnist&FMnist  &  Mnist&FMnist  &  Mnist&FMnist  &  Mnist&FMnist  &  Mnist&FMnist  &  Mnist&FMnist  \\
\cline{1-13}
{$IM1\downarrow$}           &   {$0.72$}&{$0.77$}  &  {$1.68$}&{$1.63$}  &  {$1.44$}&{$1.24$}  &  {$\bb{0.70}$}&{$\bb{0.74}$}  &  {$0.47$}&{$0.58$}  &  {$1.02$}&{$1.34$}\\ 

{$IM2\times 10\downarrow$}  &   {$0.43$}&{$0.14$}  &  {$1.08$}&{$0.26$}  &  {$1.38$}&{$0.37$}  &  {$\bb{0.36}$}&{$\bb{0.093}$}  &  {$0.29$}&{$0.12$}  &  {$2.10$}&{$0.365$}\\ 

{$FID\downarrow$}           &   {$41.12$}&{${76.52}$}  &  {$50.03$}&{$96.87$}  &  {$47.53$}&{$83.77$}  &  {$\bb{22.83}$}&{$\bb{62.31}$}  &  {$8.37$}&{$16.26$}  &  {$203.07$}&{$140.28$} \\

{$KID\times1\mathrm{e}{3}\downarrow$}  &   {$37.27$}&{${70.44}$}  &  {$44.88$}&{$91.71$}  &  {$37.24$}&{$72.71$}  &  {$\bb{13.39}$}&{$\textbf{52.71}$}  &  {$0.34$}&{$0.03$}  &  {$283.50$}&{$157.27$} 




\end{tabular}
}
\end{small}
\end{center}
\vskip -0.2in
\end{table*}

\begin{table*}[t!]
\caption{\small \textbf{Proximity and Validity}. This table compares the CF explanations methods in terms of validity and proximity. Methods that obtain high validity and low proximity are desirable.}
\label{table:metrics}
\begin{center}
\vskip -0.2in
\begin{small}
\resizebox{0.75\textwidth}{!}{%
\begin{tabular}{l||cc|cc|cc|cc||cc}

{Methods} & 
\multicolumn{2}{c|}{ExpGAN\cite{samangouei2018explaingan}} & 
\multicolumn{2}{c|}{CEM\cite{dhurandhar2018explanations}} &
\multicolumn{2}{c|}{CVE\cite{goyal2019counterfactual}}& 
\multicolumn{2}{c||}{C3LT (ours)}&
\multicolumn{2}{c}{Adv. Attack \cite{madry2018towards}}\\

\cline{2-11}
{}                 &  Mnist&FMnist  &  Mnist&FMnist  &  Mnist&FMnist  &  Mnist&FMnist  &  Mnist&FMnist \\
\cline{1-11}

{$Prox\downarrow$}  &  {$0.074$}&{$0.135$}  &  {$\bb{0.016}$}&{$\bb{0.013}$}  &  {$0.055$}&{$0.054$}  &  {$0.072$}&{$0.116$}  & {$0.229$}&{$0.196$}  \\ 

\cline{1-11}

{$Val\uparrow$}     &  {$0.997$}&{$0.998$}  &  {$0.469$}&{$0.620$}  &  {$0.231$}&{$0.145$}  &  {$\bb{0.999}$}&{$\bb{1.0}$}  &  {$0.998$}&{$1.0$}

\end{tabular}
}
\end{small}
\end{center}
\vskip -0.2in
\end{table*}

The COUT metric measures the amount of change that would be needed to move a query image into the CF class. Besides the classification change, it measures how fast the output score for the CF class maximizes, and in an opposite way, the output score for the query class minimizes. This favors methods that find sparse changes over the input features that crucially shift the output of the classifier from the query class to the CF one. In other words, COUT measures both properties \ref{pr:val} and \ref{pr:spr} as defined in the introduction.

Table \ref{table:cout} summarizes the COUT results obtained from our method and the baselines. CVE is generating the CF examples by a few discrete edits. For a fair comparison, we calculate the COUT metric for CVE slightly differently where we calculate the AUPC by measuring the output score after each edit. Both CEM and CVE perform poorly on the COUT metric as their optimization does not reach a high output score for the CF class. This also results in low validity of their generated CFs (see section \ref{sub:val}). Our method consistently outperforms the baselines in terms of the $\textit{AUPC}_{c}$, $\textit{AUPC}_{c'}$, and COUT on both MNIST and Fashion-MNIST datasets.






\vspace{-0.15in}
\subsubsection{Realism of the CFs}\label{sub:real}
\vspace{-0.05in}
In order for the generated CF examples to be \textit{relevant} as a means of explanation, they should have high \textit{realism} (property \ref{pr:real}), \ie, lie close to the data manifold of the CF class. As mentioned earlier, this is one of the main challenges for CF explanations, particularly in high-dimensional input spaces such as natural images where the pitfall of adversarial solutions becomes more prominent. To this end, we evaluate the generated CF examples from our method against the baselines in terms of their realism and how well they match the distribution of the original data. 

\cite{looveren2021interpretable} proposed \textit{IM1} and \textit{IM2} metrics that use reconstruction errors from pre-trained auto-encoders over the images from the query, CF, and all classes to assess how well the distribution of the generated CFs match the original data. A lower \textit{IM1} metric implies the CFs lie closer to the data manifold of the CF class rather than the query one. A lower \textit{IM2}, on the other hand, implies the distribution of the CFs is similar to the distribution of original data from all classes. In addition, we use Fr\'echet Inception Distance (FID) \cite{heusel2017gans} and Kernel Inception Distance (KID) \cite{binkowski2018demystifying} metrics that are well-established for evaluating the quality of the synthesized images from generative models. For all of the metrics, the lower the score, the better.

Table \ref{table:real} compares the CF explanation methods in terms of their realism. To have reference values for the aforementioned metrics, we also use images in the evaluation set from the CF class (\textit{Original Data}) as a baseline. Similar to our findings in visual comparison of the CFs, our method generates more realistic and higher quality images compared to the baselines. We also find $IM1$ and $IM2$ metrics to be more close to our visual inspection of the generated CFs where methods such as CEM, CVE, and Adv. Attacks perform very poorly. However, in terms of the FID and KID metrics, their performance is relatively better, particularly for the Fashion-MNIST dataset. 


\vspace{-0.15in}
\subsubsection{Validity}\label{sub:val}
\vspace{-0.05in}
When explaining the classifier $f$ using CF examples, the generated CFs are expected to lie within the decision boundaries of the CF class $c'$, \ie, be valid. In addition to COUT and in order to have a simple and intuitive metric to measure the validity of CFs (property \ref{pr:val}), we define the  $Val$ metric,
\vspace{-0.05in}
\begin{equation}\label{eq:valid}
    Val = \frac{1}{N}\sum_{n=1}^{N} \mathbbm{1}_{f(\vx'_n), c'}
\end{equation}
where $\mathbbm{1}_{f(\vx'_n), c'}$ is the indicator function that the prediction of the classifier $f$ for the $n$-th CF example $\vx'_n$ is $c'$, and $N$ is the total number of CFs. This measures the fraction of the generated CF examples that are correctly predicted by the classifier $f$ to the CF class $c'$. Table. \ref{table:metrics} shows the obtained results from our method and compares it against baselines. Our method achieves very high $Val$ on both MNIST and Fashion-MNIST datasets. However, \cite{dhurandhar2018explanations, goyal2019counterfactual} struggle to generate valid examples, hence their explanations have low \textit{faithfulness}. High $Val$ scores obtained from the Adv. Attack also point out that sole reliance on the validity for CF evaluation can be misleading and it should always be considered along with other evaluation criteria such as COUT or realism (see section \ref{sub:real}).

\vspace{-0.15in}
\subsubsection{Proximity}\label{sec:prox}
\vspace{-0.05in}
In order to generate CF examples, minimal changes to the features of the query image are favorable (property \ref{pr:prx}). We simply define the \textit{proximity} metric as the mean of feature-wise $L1$ distances between the query and CF examples,
\begin{equation}\label{eq:prox}
    Prox = \sum_{n=1}^{N}\frac{||\vx_n-\vx_n'||_1}{NCHW} 
\end{equation}
where $\vx_n$ and $\vx'_n$ are the $n$-th query and CF example from the evaluation set, and $C$, $H$ and $W$ are the number of channels, height, and width of the input image, respectively. Table \ref{table:metrics} compares our method against the baselines in terms of proximity of the generated CFs. It can be seen that \cite{dhurandhar2018explanations, goyal2019counterfactual} outperform other methods on this metric. However, as mentioned earlier, the generated CFs from their method do not have high realism (see section \ref{sub:real}) and faithfulness (see section \ref{sub:val}) hence not reliable. 

\vspace{-0.05in}
\section{Conclusion}
\vspace{-0.05in}
{
In this paper, we presented a novel framework for generating counterfactual explanations by
learning a transformation function in the latent space of a generative model (GAN/VAE) with a combination of several loss functions including cycle-consistency. Extensive experiments show that our approach outperforms prior work across all metrics, as well as possessing two desirable properties: \textit{First}, it does not require joint training with a generative model hence can be plugged into state-of-the-art generative algorithms to generate high-resolution CFs. \textit{Second}, once learned, our approach can generate CF examples on-the-fly during inference time which makes it ideal to be used in practical systems to explain deep networks.
}

\vspace{-0.1in}
\section*{Acknowledgements}
\vspace{-0.05in}
This work is supported in part by DARPA contract N66001-17-2-4030. We also thank Dr. Xin Wang from the Albany Samaritan hospital who inspired us to perform this research.




{\small
\bibliographystyle{files/ieee_fullname}
\bibliography{ref}
}

\clearpage
\appendix
\section*{Supplementary Materials}

\section{Choice of Hyperparameters}\label{sec:hyperparams}

There are no implementations available online for ExplainGAN\cite{samangouei2018explaingan} so we had to implement it on our own. The set of hyperparameters are also not stated in the paper. We found the best set of hyperparameters though cross-validation. We weighted each loss term ($\gL_\text{classifier}$, $\gL_\text{recon}$, and $\gL_\text{prior}$) in the objective equally with coefficient $1$. Their prior loss is also comprised of multiple loss terms ($\gL_\text{const}$, $\gL_\text{count}$, $\gL_\text{smoothness}$, and $\gL_\text{entropy}$). We used the coefficient $1000$ for the count loss and set the rest as $1$. In addition, for the choice of $\kappa$, which controls the effect of count loss, we used $0.05$ and $0.1$ for MNIST and Fashion-MNIST datasets, respectively. 

For CEM\cite{dhurandhar2018explanations}, we used the default set of hyperparameters available at the official Github repository \footnote{\url{https://github.com/IBM/Contrastive-Explanation-Method}} and set $\gamma=100$, which controls the auto-encoding error. We used the implementation provided by the authors for the CVE\cite{goyal2019counterfactual} method. Their feature replacement search occurs at the last convolutional layer of the classifier. In order to generate comparable CFs with other baselines, we slightly changed the objective in both CEM and CVE methods and set the target to a user-specified class rather than \textit{maximum-non-query} class. 

We added the PGD targeted adversarial attack \cite{madry2018towards} as a baseline. We used the \textit{torchattacks}\footnote{\url{https://github.com/Harry24k/adversarial-attacks-pytorch}} library for doing the attacks. We set the step size $\alpha=1/255$, maximum number of steps to $1000$, and  maximum step sizes of $\epsilon=64/255$ and $\epsilon=72/255$ for MNIST and Fashion-MNIST datasets, respectively.

Each loss term in the main objective of C3LT is scaled by a coefficient which the values are obtained through cross-validation:
\begin{equation}
     \gL_{c3lt} = \gL_{cls} + \alpha \gL_{prx} + \beta\gL_{cyc} + \gamma\gL_{adv}
\end{equation}
where we set $\{\alpha=0.1, \beta=0.1, \gamma=0.001\}$ for both the MNIST and Fashion-MNIST datasets. In addition, we found one non-linear step to be sufficient in our experiments and set $n=1$.

We used the PyTorch\cite{NEURIPS2019_9015} framework to implement and evaluate all methods (including C3LT) and deep neural networs, except for the CEM which we used the original implementation in Tensorflow\cite{tensorflow2015-whitepaper}. Across all methods, we used the same pretrained classifier for both MNIST ($99.4\%$ accuracy) and Fashion-MNIST ($91.5\%$) datasets. 



\section{Computation Time Comparison}\label{sec:time_comp}

It is ideal that the CF examples are generated on the fly. This is particularly helpful when users and machine explanations interact. Many approaches, including CEM and CVE, generate CF explanations by solving iterative optimization problems and there are no training phase. Hence, it is not a surprise such methods are not \textit{fast} and cannot be used for real-time applications. On the other hand, our method generate CFs orders of magnitude faster than iterative methods. This is mainly due to the fact that our method only does a forward pass in the C3LT pipeline during inference time. We present the average computation time (per sample) to generate CF explanations from our method and baselines for the MNIST dataset in Table. \ref{table:time}. ExpGAN is showing comparable results since their approach also does a forward pass at inference time. However, it is slightly slower as their generator has multiple heads while ours does not. We used the same batch size of 256 for evaluating our method and ExpGAN. CVE and CEM generate explanations one sample at a time. To run the experiments, we used a HP Z640 Workstation with a single NVIDIA GeForce RTX 2080-Ti GPU.  

\begin{table*}[thb!]
\caption{\textbf{Computation Time Comparison of CFs}. This table shows the average computation time (per sample) to generate CF examples in seconds for the MNIST dataset. C3LT is relatively faster than ExpGAN while being significantly faster than CVE and CEM --- as they solve iterative optimization problems to generate CFs. At inference time, our method only does a forward pass through the C3LT pipeline to generate CFs. This makes our method suitable for CF explanation generation on the fly.}
\label{table:time}
\begin{center}
\begin{small}
\resizebox{0.6\textwidth}{!}{%
\begin{tabular}{l||c|c|c|c}

{Methods} & 
\multicolumn{1}{c|}{ExpGAN\cite{samangouei2018explaingan}} & 
\multicolumn{1}{c|}{CEM\cite{dhurandhar2018explanations}} &
\multicolumn{1}{c|}{CVE\cite{goyal2019counterfactual}}& 
\multicolumn{1}{c}{C3LT (ours)}\\

\cline{2-5}
\cline{1-5}

{Time (\textit{sec})$\downarrow$}  &  {$1.28\mathrm{e}{-5}$}  &  {${68.34}$}  &  {$2.91\mathrm{e}{-2}$}  &  {$\bb{9.23\mathrm{e}{-6}}$}  \\

\end{tabular}
}
\end{small}
\end{center}
\end{table*}

\section{Ablation Study}\label{sec:ablation}

To analyze the contribution of each loss term in $\gL_{c3lt}$, we conduct an ablation study. We respectively add the $\gL_{prx}$, $\gL_{cyc}$, and $\gL_{adv}$ loss terms to the $\gL_{cls}$ and evaluate the generated CFs in terms of the metrics explained in the paper. Table \ref{table:ablation} shows the obtained results from our ablation study on MNIST dataset. Minimizing the $\gL_{cls}$ generates images that are in the CF class. However, it does not consider minimal perturbations to the input in order to change the decision of the classifier, \ie, the input images and the obtained CFs are distant. This is reflected in the proximity metric ($Prox$). Adding the $\gL_{prx}$ encourages such minimal changes and improves the proximity score while maintaining almost perfect validity. As one can expect, this improves the COUT metric as well. Adding the $\gL_{cyc}$ further regularizes the training and helps with learning more accurate transformations. This further improves the proximity and COUT metrics. Finally, adding the $\gL_{adv}$ helps with improving the realism metrics (IM1, IM2, FID, and KID). This ensures the generated CFs stay close to the data manifold, resulting in changes that are actionable and sensible to humans. 

\begin{table*}[thb!]
\caption{\textbf{Ablation Study}. In this table, we provide the quantitative results obtained from the ablation study of the C3LT objective. We respectively add the $\gL_{prx}$, $\gL_{cyc}$, and $\gL_{adv}$ loss terms to the $\gL_{cls}$ and evaluate the generated CFs using the CF evaluation metrics.}
\label{table:ablation}
\begin{center}
\begin{small}
\resizebox{0.95\textwidth}{!}{%
\begin{tabular}{l||c|c|c|c|c|c|c}

{Lost Terms} & 
\multicolumn{1}{c|}{$COUT \uparrow$} & 
\multicolumn{1}{c|}{$IM1\downarrow$} &
\multicolumn{1}{c|}{$IM2\times 10\downarrow$} &
\multicolumn{1}{c|}{$FID\downarrow$} &
\multicolumn{1}{c|}{$KID\times1\mathrm{e}{3}\downarrow$}& 
\multicolumn{1}{c|}{$Prox\downarrow$}&
\multicolumn{1}{c}{$Val\uparrow$}\\
\cline{1-8}
{$\gL_{cls}$}  &  {${0.897}$}  &  {${0.47}$}  &  {${0.29}$}  &  {${40.84}$} &  {${29.22}$} & {${0.113}$} & {$\bb{1.0}$} \\

{$\gL_{cls}+\gL_{prox}$}  &  {${0.935}$}  &  {${0.65}$}  &  {${0.34}$}  &  {${33.40}$} &  {${22.36}$} & {${0.076}$} & {${0.998}$} \\

{$\gL_{cls}+\gL_{prox}+\gL_{cyc}$}  &  {${0.943}$}  &  {${0.78}$}  &  {${0.38}$}  &  {${29.95}$} &  {${18.65}$} & {$\bb{0.069}$} & {${0.998}$} \\

{$\gL_{cls}+\gL_{prox}+\gL_{cyc}+\gL_{adv}$ (C3LT)}  &  {$\bb{0.948}$}  &  {$\bb{0.70}$}  &  {$\bb{0.36}$}  &  {$\bb{22.83}$} &  {$\bb{13.39}$} & {${ 0.072}$} & {${0.999}$} \\

\end{tabular}
}
\end{small}
\end{center}
\end{table*}

\section{Distinction from CycleGAN}\label{sec:cyclegan}

Here, we first evaluated our method against CycleGAN \cite{zhu2017unpaired}. Then, we elaborate on the similarities of C3LT and CycleGAN and how they distinct from each other. Finally, we showcase debugging a classifier using C3LT where methods such as CycleGAN are not useful.

\begin{figure*}[hbt!]
\begin{center}
{ 
\includegraphics[width=0.95 \linewidth]{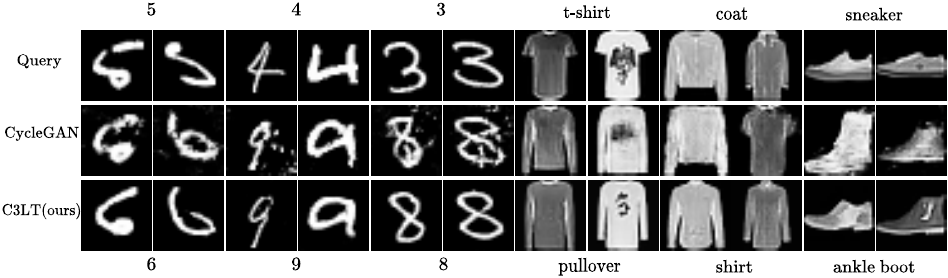} 
}
\end{center}
    \caption{\textbf{Visual comparison of C3LT and CycleGAN}. This figure compares the CF examples generated by C3LT and CycleGAN on MNIST and Fashion-MNIST datasets. CF examples obtained from CycleGAN are often blurry with scattered perturbation with respect to the query image while the CF images from C3LT are more realistic, sharp, and close to the original query image.} \label{fig:cycle}
\end{figure*}

CycleGAN learns image-to-image translation using generative adversarial training. We used the images from the query and CF classes as the input and output image domains. We used the official implementation of CycleGAN\footnote{\url{https://github.com/junyanz/CycleGAN}} and trained translation functions on MNIST and Fashion-MNIST datasets. Fig. \ref{fig:cycle} visually compares the CF examples obtained from CycleGAN and C3LT on both datasets and various pairs. The generated CFs from C3LT are more realistic and sharp while CycleGAN results are often blurry and scattered with meaningless perturbations across the image (e.g. 3 to 8). In addition, the C3LT translations are more proximal to the original input image (e.g. sneaker to boot). 

Table \ref{table:cycle} shows quantitative comparison of CycleGAN and C3LT in terms of CF metrics on MNIST dataset. Following the insights obtained from the visual comparison of the CFs in Fig. \ref{fig:cycle}, Table \ref{table:cycle} corroborates that the CF examples from C3LT are more realistic and have higher quality. In addition, the COUT score obtained in this comparison shows that the C3LT generates more valid and sparse explanations. 

\begin{table*}[t!]
\caption{\textbf{Quantitative Comparison of C3LT against CycleGAN}. We compare the CF obtained from CycleGAN and our method in terms of CF metrics. CF examples generated from C3LT are more realistic and have higher quality.In addition, the COUT score obtained in this experiment shows that the C3LT generates more valid and sparse explanations than CycleGAN.}
\begin{center}
\label{table:cycle}
\begin{small}
    \resizebox{0.85\textwidth}{!}{%
        \begin{tabular}{l||c|c|c|c|c|c|c}
            {{\diagbox[width=10em]{Method}{Metric}}} & 
            \multicolumn{1}{c|}{$COUT \uparrow$} & 
            \multicolumn{1}{c|}{$IM1\downarrow$} &
            \multicolumn{1}{c|}{$IM2\times 10\downarrow$} &
            \multicolumn{1}{c|}{$FID\downarrow$} &
            \multicolumn{1}{c|}{$KID\times1\mathrm{e}{3}\downarrow$}& 
            \multicolumn{1}{c|}{$Prox\downarrow$}&
            \multicolumn{1}{c}{$Val\uparrow$}\\
            \cline{1-8}
            {C3LT(ours)}  &  {$\bb{0.948}$}  &  {$\bb{0.70}$}  &  {$\bb{0.36}$}  &  {$\bb{22.83}$} &  {$\bb{13.39}$} & {$\bb{ 0.072}$} & {$\bb{0.999}$} \\
            {Cycle-GAN\cite{zhu2017unpaired}}  &  {${0.894}$}  &  {$0.716$}  &  {$0.49$}  &  {$43.4$} &  {$41.61$} & {$ 0.089$} & {$0.988$}
        \end{tabular}
    }
\end{small}
\end{center}
\end{table*}

CycleGAN \cite{zhu2017unpaired} learns unpaired image translation using a cycle-consistent generative adversarial training. While the cycle-consistency in the C3LT is inspired by it, there are two main differences that separates our work. First, that our cycle-consistency is in the latent space of a given (pre-trained) generator and the transformations are occurring in the latent space, rather than direct image-to-image translation. This is favorable as our method can be easily plugged into state-of-the-art pretrained generative models (GANs, VAEs, etc.) and discard training them from scratch. Second, the CycleGAN is \textit{not} explaining a classifier. Indeed, CycleGAN uses two different discriminators to keep the translated images close to the data manifold of each target class. However, the main goal of this paper is to explain a given classifier through CF explanations. In the follwoing, we show that C3LT can be used for debugging a \textit{faulty} classifier while methods such as CycleGAN are not helpful. 

\begin{figure*}[hbt!]
\begin{center}
\subfloat[]{\includegraphics[width=0.4\linewidth]{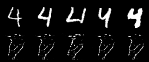}
}
\subfloat[]{\includegraphics[width=0.4\linewidth]{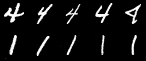}
}
\end{center}

\caption{\textbf{Debugging a {faulty} Classifier}. Here, we showcase the capability of C3LT in debugging a \textit{rigged} classifier while methods such as CycleGAN are not helpful. We choose images from class 4 as the query images (top row) and set the a) \textit{left-out} b)\textit{non-left-out} class as the target for the CF examples (bottom row). While the CFs for the left-out class are not interpretable, the CFs for the non-left-out class are intuitive. GAN-type approaches such as CycleGAN do not interpret different classifiers and would generate regular digits $9$ and $1$ in either case, respectively.}
\vskip -0.05in 
\label{fig:debug}
\end{figure*}

We use C3LT to provide explanations for a \textit{faulty} classifier. Here, we simply rig a classifier by depriving it from seeing examples from a specific class during the training. Put it differently, we train a classifier that lacks knowledge regarding a specific class while it is having a reasonably well performance on the rest the classes. To that end, we trained a classifier (identical to the one used for experiments) on MNIST dataset while discarding the training examples from class 9. This classifier obtains $\sim 89\%$ test accuracy --- only missing test samples from the \textit{left-out} class 9. We then attempt to explain the decision of this classifier using C3LT as shown in Fig. \ref{fig:debug}. Choosing the class 4 for the query images, we set classes 9 (left-out class) and 1 (non-left-out class) as the target for the CF explanations. As one might expect, the CFs for the left-out class are not interpretable and meaningful which emphasizes the classifier lacks knowledge regarding the target class. On the other hand, when choosing the non-left-out class 1, the CF explanations are intuitive and might be helpful to a user, whereas GAN-type approaches such as CycleGAN will just continue to generate normal digits without using the classifier. This is a simple scenario for understanding the weakness of a classifier; however, it emphasises the substantial differences between C3LT and methods such as CycleGAN. While our method can explain \textit{any} classifier, CycleGAN and other GAN-type methods are not of use.

\section{Traversal in the Latent Space}\label{sec:traversal}

\begin{figure*}[ht]
\vskip -0.1in
  \centering
  \includegraphics[width=0.75 \linewidth]{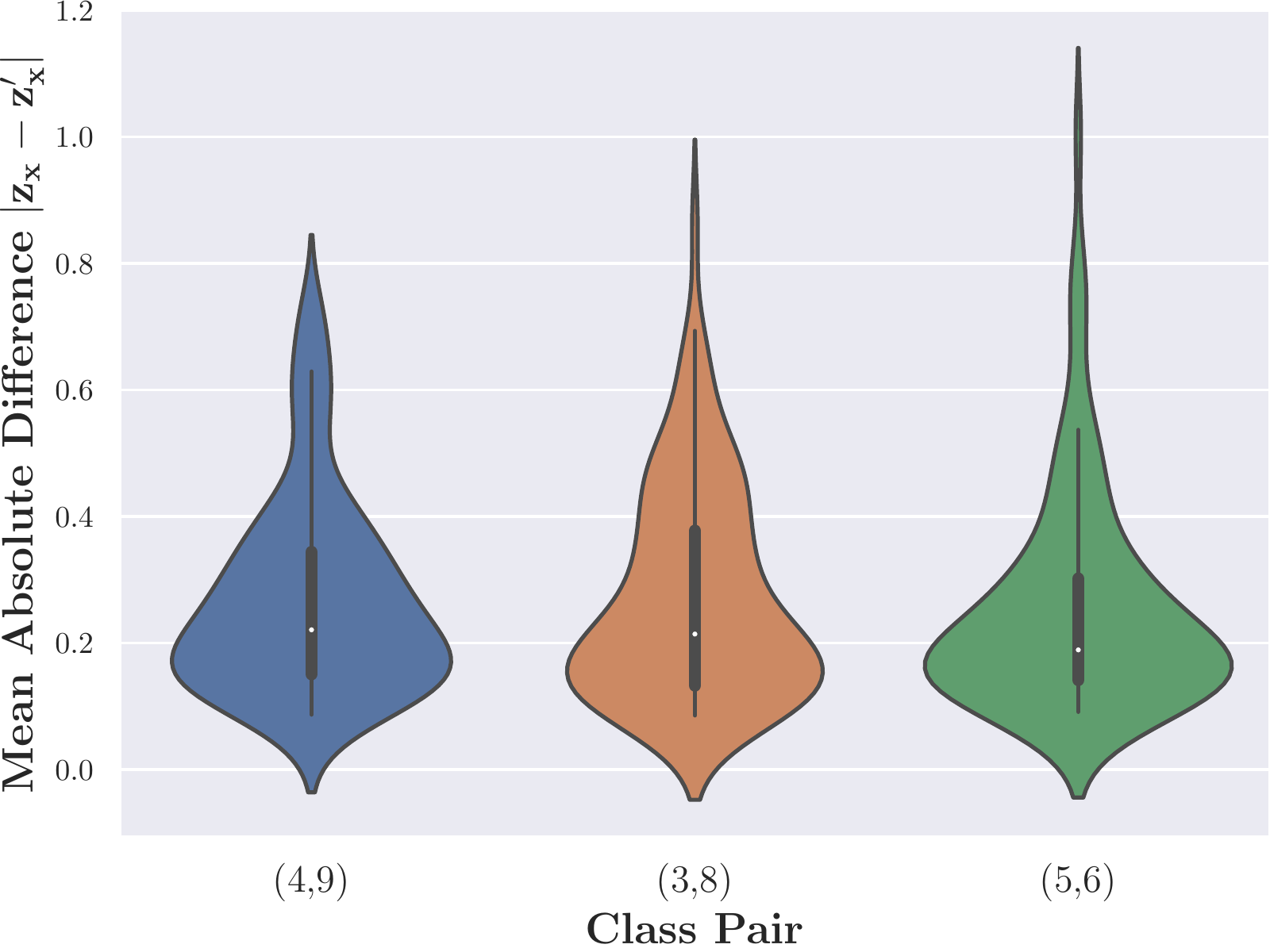}
  \vskip -0.05in
  \caption{Mean absolute difference of the input and CF latent codes using C3LT for class pairs from MNIST. It can be observed the that the learned transformations are not sparse.}
  \label{fig:box}
\vskip -0.1in
\end{figure*}

\textbf{"Does C3LT lead to disentangled transformations in the latent space?"} Normal VAE/GANs do not generate disentangled latents. Fig. \ref{fig:box} is a violin plot showing the mean absolute difference in latent dimensions between the original and CF images, for three class pairs from MNIST. It shows most latent dimensions are changed (with average magnitude of 0.2). Note our main goal is to build a method to generate realistic and high-resolution CF images for explaining classifiers, so sparsity of latent traversal is interesting future work but orthogonal to this goal.

\begin{figure*}[ht]
\vskip -0.1in
  \centering
  \includegraphics[width=0.85 \linewidth]{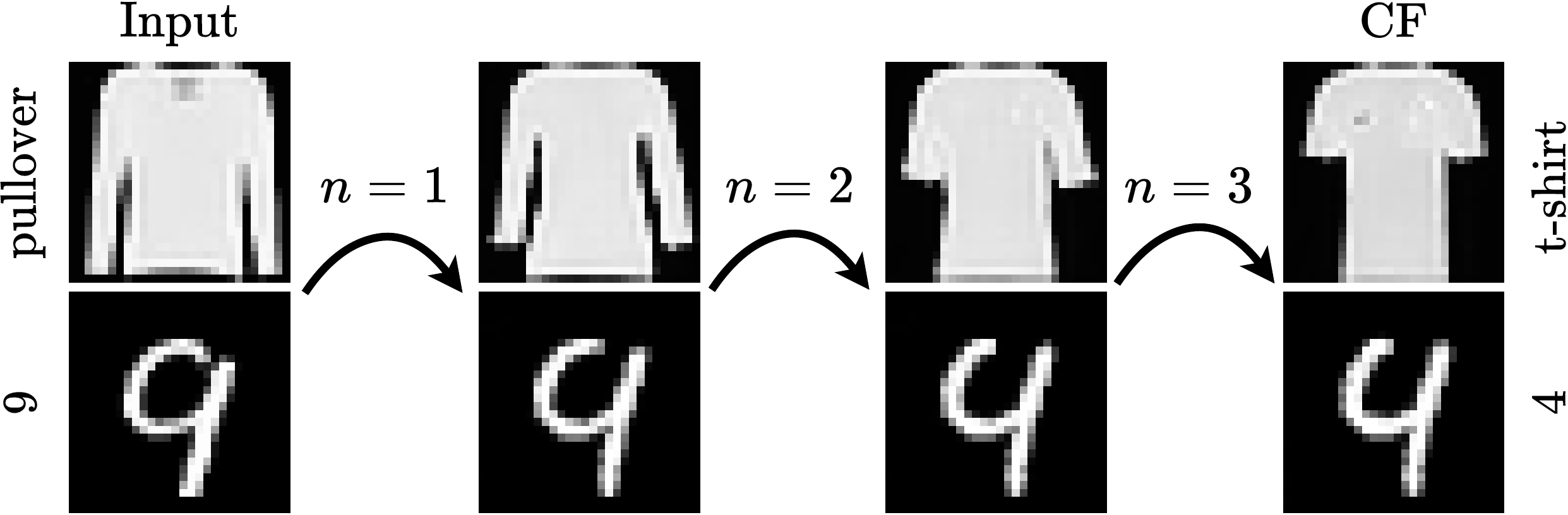}
  \vskip -0.05in
  \caption{Traversal with $n=3$ steps in the latent space of the generator going from the input to the CF example.}
  \label{fig:trv}
  \vskip -0.1in
\end{figure*}

However, when steering from the input to the CF in the latent space, we observe meaningful traversal. Fig. \ref{fig:trv} illustrates this for (pullover, t-shirt) and ($4$, $9$) class pairs with $n=3$ discrete steps (see Eq.\ref{eq:cfl_g} in the paper).

\section{Additional Visual Examples}\label{sec:more_visual}

In the following, we show more CFs generated from our method and baselines for both MNIST and Fashion-MNIST datasets.

\begin{figure*}[hbt!]
\begin{center}
{ 
\includegraphics[width=1 \linewidth]{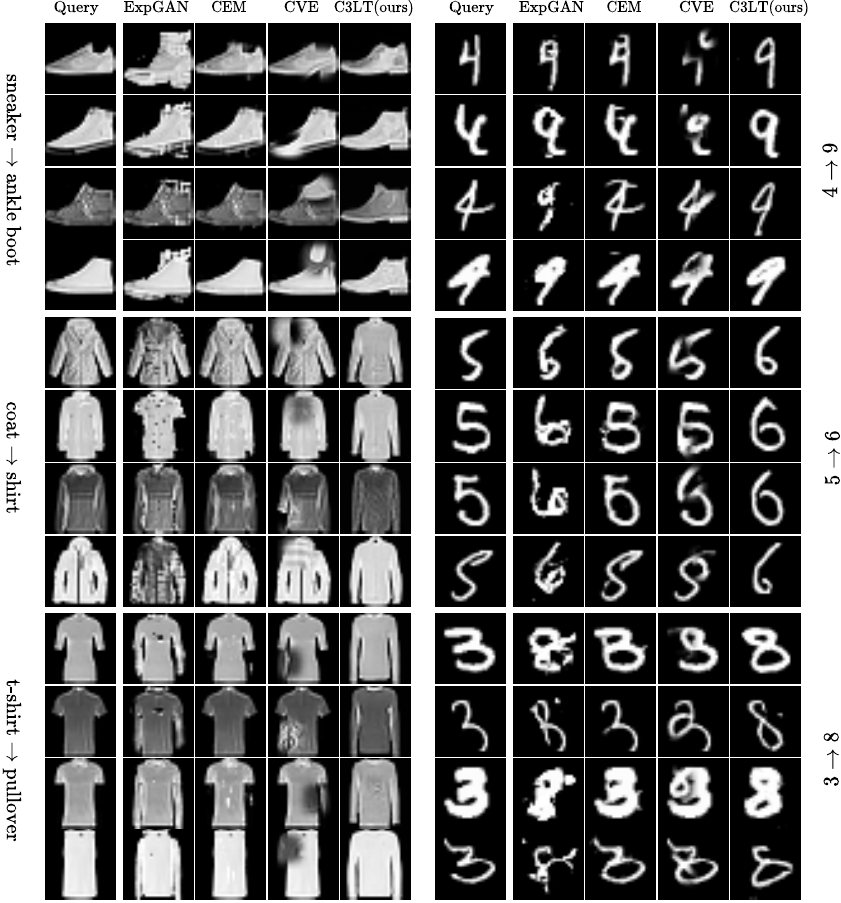} 
}
\end{center}
    \caption{\textbf{Visual Comparison of the CFs}. This figure illustrated the generated CFs from C3LT, ExplainGAN, CEM, and CVE for both MNIST and Fashion-MNIST datasets. For MNIST, we show the (3,8), (4,9), and (5,6) pairs. For Fashion-MNIST, the pairs are (t-shirt, pullover), (coat, shirt), and (sneaker, ankle boot). It can be noted thatthe generated CFs from the CEM and CVE to be adversarial and off the data manifold. Compared to the ExpGAN, the generated CFs from our method are consistently more realistic and interpretable.} \label{fig:more_nist}
\end{figure*}

\end{document}